\pdfoutput=1

\documentclass[11pt]{article}

\usepackage[final]{emnlp2023}

\usepackage{times}
\usepackage{latexsym}

\usepackage[T1]{fontenc}

\usepackage[utf8]{inputenc}
\usepackage{csquotes}

\usepackage{microtype}

\usepackage{inconsolata}

\usepackage{amsmath}
\usepackage{booktabs}
\usepackage{graphics}
\usepackage{changepage}
\usepackage{graphicx, multirow}
\usepackage{amsmath}
\usepackage[linesnumbered,ruled]{algorithm2e}
\usepackage{makecell}
\usepackage{array}
\usepackage{arydshln} 
\usepackage[toc,page]{appendix}
\usepackage[export]{adjustbox}
\usepackage{tabularx}
\usepackage{tikz}
\usepackage{enumitem}
\newcommand*\mycirc[1]{%
\begin{tikzpicture}[baseline=(C.base)]
\node[draw,circle,inner sep=1pt,minimum size=.5ex](C) {#1};
\end{tikzpicture}}

\usepackage{enumitem}

\definecolor{deeppink}{rgb}{1.0, 0.08, 0.58}

\newcommand{\ourdataset}{MCPDial}

\definecolor{cadmiumgreen}{rgb}{0.0, 0.42, 0.24}

\title{ 
MCPDial: A Minecraft Persona-driven Dialogue Dataset
}

\author{Seyed Hossein Alavi$^{\thanks{This work was partially completed during the author’s internship at Microsoft.}~1,2}$~~Sudha Rao$^3$ ~~Ashutosh Adhikari$^3$\\
~~\textbf{Gabriel DesGarennes1}$^{3}$~~\textbf{AkankshaMalhotra}$^3$~~\textbf{Chris Brockett}$^3$~~\textbf{Mahmoud Adada}$^3$\\~~\textbf{Raymond Ng}$^{1}$~~\textbf{Vered Shwartz}$^{1,2}$~~\textbf{Bill Dolan}$^{3}$\\
$^1$ University of British Columbia~~
$^2$ Vector Institute for AI~~
$^3$ Microsoft Research\\
{\tt salavis@cs.ubc.ca \tt sudha.rao@microsoft.com}}

\begin{document}
\maketitle

\begin{abstract}
We propose a novel approach that uses large language models (LLMs) to generate persona-driven conversations between Players and Non-Player Characters (NPC) in games. Showcasing the application of our methodology, we introduce the Minecraft Persona-driven Dialogue dataset (\ourdataset). Starting with a small seed of expert-written conversations, we employ our method to generate hundreds of additional conversations. Each conversation in the dataset includes rich character descriptions of the player and NPC. The conversations are long, allowing for in-depth and extensive interactions between the player and NPC. \mbox{\ourdataset} extends beyond basic conversations by incorporating canonical function calls (e.g. ``Call find a resource on iron ore'') between the utterances. Finally, we conduct a qualitative analysis of the dataset to assess its quality and characteristics.
\end{abstract}

\section{Introduction}
\label{sec:introduction}

Natural language generation vastly improved in recent years thanks to large language models \cite[LLMs;][]{radford2019language,NEURIPS2020_1457c0d6}, benefiting NLP applications across the board, such as text summarization, machine translation, and dialogue systems. Trained on web texts, LLMs are proficient in generating generic texts about various topics, but they lack a consistent ``personality''. Many applications can benefit from tailoring the output to the user's intent \cite{dudy-etal-2021-refocusing}, cultural background \cite{hershcovich-etal-2022-challenges}, gender \cite{rabinovich-etal-2017-personalized}, and more. 

In this paper, we focus on persona-driven dialogue generation, where the generated conversation is consistent with a given character description \cite{li-etal-2016-persona,zehngetal2020}. 
Persona-driven dialogue generation is especially relevant for video games, where players interact with Non-Player Characters (NPCs) that have different roles, backgrounds, and motivations. Nevertheless, the availability of high-quality dialogue datasets with a persona-driven focus is currently limited, especially within the context of video games.

\begin{figure}[t]
    \centering
    \includegraphics[frame,width=0.48\textwidth]{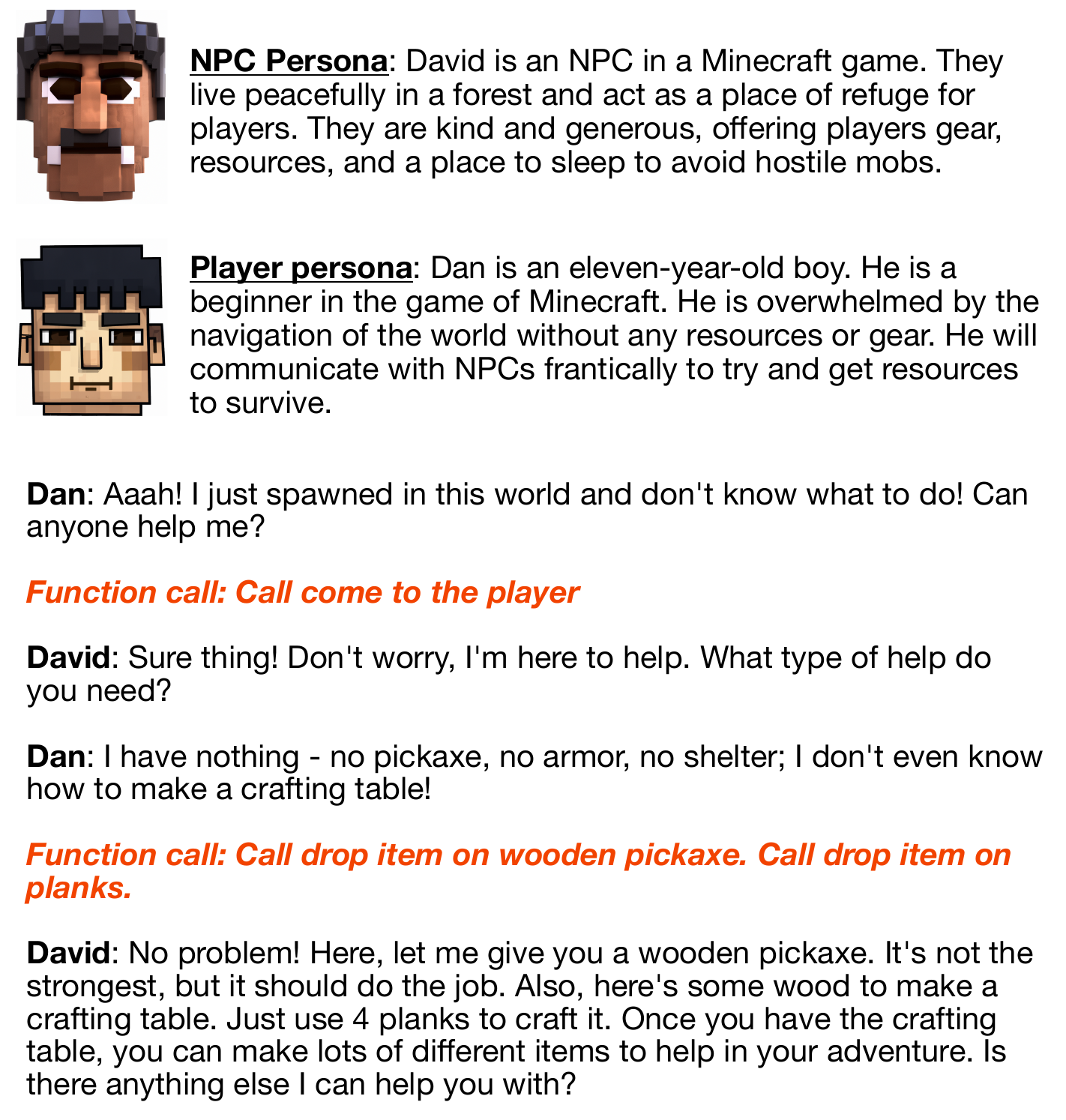}
    \caption{Example persona descriptions and the beginning of the automatically generated conversation from our dataset.}
    \vspace{-5pt}
    \label{fig:example_conversation}
\end{figure}

We propose a novel approach that uses LLMs to generate persona-driven conversations between Players and NPCs in games. Our approach leverages the power of LLMs to generate fluent and coherent conversations that follow the persona descriptions, while also incorporating game-specific function calls to increase interactivity. 

To demonstrate the effectiveness and versatility of our approach, we apply it to Minecraft, a popular sandbox video game \cite{persson2011minecraft}, and introduce the \ourdataset{} dataset (\textbf{M}ine\textbf{c}raft \textbf{P}ersona-driven \textbf{Dial}ogue). Figure~\ref{fig:example_conversation} shows an example from our dataset. Human evaluation shows that \ourdataset{} consists of high-quality conversations in terms of fluency, persona consistency, and function call accuracy. 

\ourdataset{} serves as a valuable resource for training and evaluating conversational agents within the Minecraft domain, enabling the exploration of nuanced exchanges and the development of more immersive and context-aware gaming experiences.

\section{Background}
\label{sec:background}

\subsection{Persona-Driven Dialogue Generation}
\label{sec:background:persona_driven_dialogue}

Previous work on persona-driven dialogue generation has mainly focused on two aspects: how to model the personas of the dialogue agents and how to generate responses that are consistent with the persona. For the first aspect, earlier work learned persona embeddings from the dialogue history \cite{li-etal-2016-persona}, while more recent work leverages explicit persona descriptions  \cite{qianetal2018,majumder-etal-2020-like,huang2023personalized}. With respect to the second aspect, persona descriptions may be used to verify the consistency of the generated response \cite{song-etal-2021-bob}, or incorporated into the model and then conditioned on, with a neural memory \cite{xuetal2022}, or encoded using a separate encoder \cite{huang2023personalized}.

The recent success of LLMs in persona-driven dialogue generation is mostly on simple conversations, while in this work, we address the challenge of generating persona-driven dialogues in games. Games provide complex and interactive scenarios where the dialogue agents need to perform various tasks and functions besides chatting.

\subsection{Persona Focused Datasets} 
\label{sec:background:persona_focused_dataset}

Previous research efforts focused on datasets for open-ended persona-driven dialogues (``chit chat''). PersonaChat \cite{zhang-etal-2018-personalizing} consists of chit chat conversations between 1k different personas. The dataset was the basis for the ConvAI2 NeurIPS competition \cite{DBLP:journals/corr/abs-1902-00098}. \newcite{mazare-etal-2018-training} released a similar, larger-scale dataset extracted from Reddit, covering 5 million personas. 
MSC \cite{xu-etal-2022-beyond} consists of multiple chat sessions between crowdsourcing workers, where participants learn about each other's interests and discuss knowledge acquired from past sessions. They showed that existing models trained on short conversations are limited in their performance in long conversations.

Our dataset \ourdataset{} similarly focuses on long conversations. In addition, differently from most prior work, the conversations in \ourdataset{} are not open ended, but instead are grounded in a game environment.

The LIGHT dataset \cite{urbanek-etal-2019-learning} is the most closely related dataset to ours, focusing on a fantasy text adventure game. LIGHT was collected by crowdsourcing conversations grounded in an environment described by location, objects, and character personas. In this paper, we show that conversations grounded in the rich environment of Minecraft can be generated by LLMs with minimal supervision.

\begin{table}[t]
    \centering
    \small
    \begin{tabularx}{\columnwidth}{Xl}
        \toprule
        \textbf{Function Call} & \textbf{Args}\\
        \midrule
         Come to the player & N/A \\
         List all items in inventory & N/A  \\
         Get crafting recipe on & iron sword, etc. \\
         Locate a place on & realms, caves, etc. \\
         Check if item is in inventory on & pickaxe, sword etc.\\
         Drop item on &potions, sword, etc.\\
         Take items from player on & diamonds, etc.\\
         \bottomrule
    \end{tabularx}
    \caption{Sample of function calls from \ourdataset{}.}
    \label{tab:sample_function_calls}
\end{table}

\section{\ourdataset\footnote{Dataset is available at: \url{https://github.com/salavi/MCPDial-A-Minecraft-Persona-driven-Dialogue-Dataset}}}
\label{sec:MCPDial}

We present the creation process of \ourdataset. We first collected a small number of gold-standard persona descriptions and conversations (Section~\ref{sec:gold_collection}). Then, we augmented the data by generating additional instances using LLMs (Section~\ref{sec:framework}).

\subsection{Gold Data Collection Process}
\label{sec:gold_collection}

\paragraph{Persona Descriptions.} We collected 250 Minecraft NPC persona descriptions from trained annotators. These personas offer a more comprehensive description beyond mere attributes like politeness or helpfulness. In some cases, the personas include embedded tasks. For instance, an NPC might offer a reward upon completion of a specific task. An illustrative example of such personas is provided in Figure~\ref{fig:example_conversation}.

We then tasked the annotators with crafting player personas corresponding to the provided NPC personas. These personas offer a distinct perspective, representing a player interacting with the NPC without any prior knowledge of the NPC personas. To enhance the diversity of our dataset, we collected 3 player personas for each NPC persona. The player personas include their level of proficiency in Minecraft, which is categorized into beginner, intermediate, and expert. 

\paragraph{Function Calls.} We compiled a list of 20 canonical function calls that might be employed during a conversation. Certain function calls necessitate specific arguments, such as ``call take item from player on \textbf{diamonds}'', while others don't take arguments, such as ``Call list items in the inventory''. We also include instances where multiple function calls were combined within a single utterance, allowing for more dynamic and complex interactions (Fig.~\ref{fig:example_conversation}). Table~\ref{tab:sample_function_calls} exemplifies the function calls, and Appendix~\ref{appendix.all_function_calls} provides the full list.

\paragraph{Dialogues.} We randomly sampled an NPC persona and a player persona, and instructed the annotators to compose a scene (potentially featuring environments like a seaside, desert, or cave, along with minor details), and an entire plausible conversation between the two participants. We collected 49 conversations, which serve as our gold data.

\begin{figure}[t]
    \centering
\includegraphics[width=0.85\columnwidth]{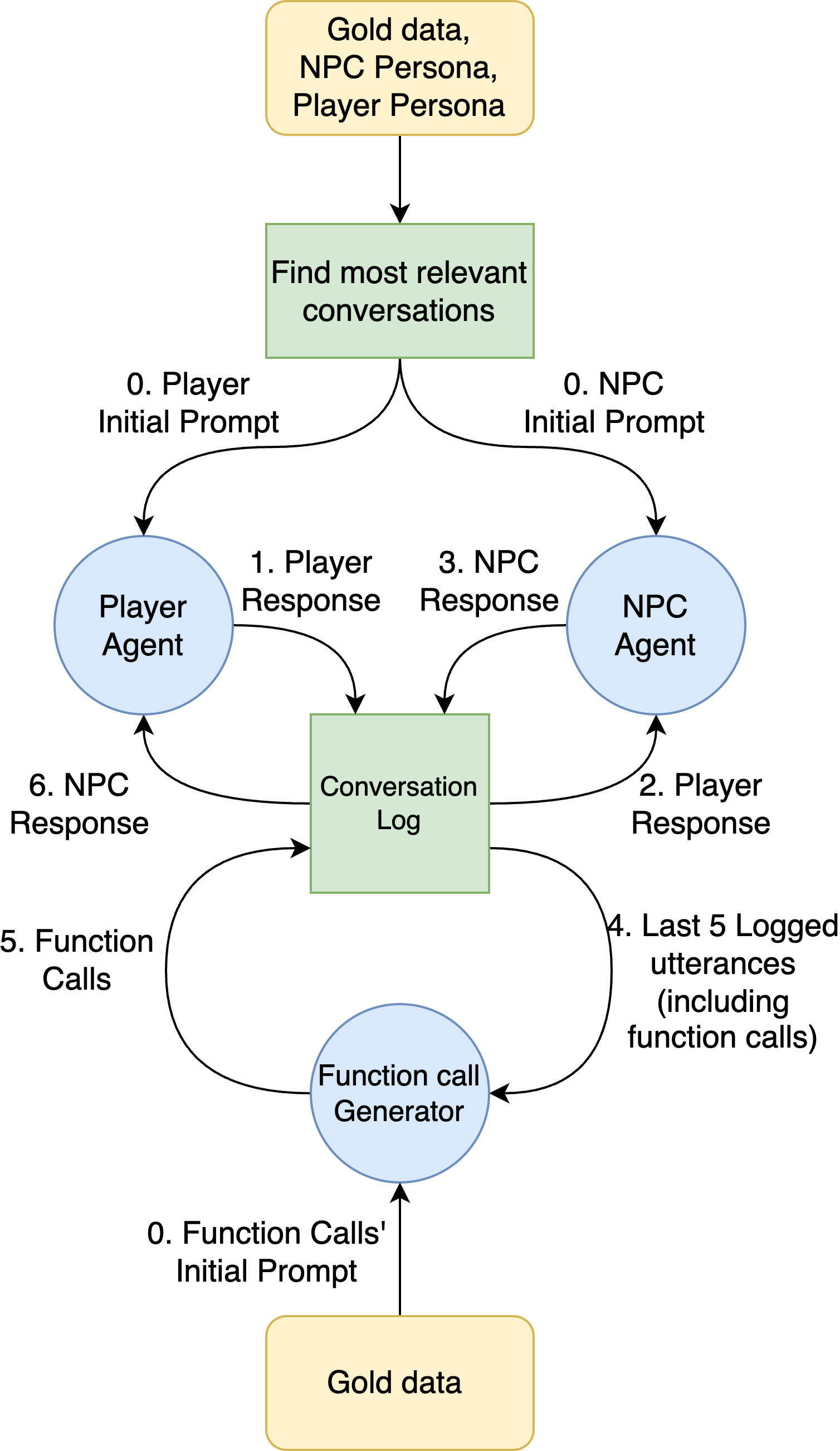}
    \caption{An overview of our conversation generation approach.
    }
    \label{fig_framework}
\end{figure}

\subsection{Conversation Generation}
\label{sec:framework}

Figure~\ref{fig_framework} provides an overview of our data augmentation method for generating high-quality persona-driven conversations between NPC and players based on their persona descriptions. The method 
consists of the following steps. For a given pair of persona descriptions, we first find the most relevant conversations from the gold dataset (Section~\ref{sec:relevant_conversations}). These conversations are used as in-context examples for the LLM which is then tasked with generating the utterances (Section~\ref{sec:utterances}) and function calls (Section~\ref{sec:function_calls_gen}). We decided to use the \texttt{GPT3.5} model for our experiments. 

\subsubsection{In Context Examples}
\label{sec:relevant_conversations}

To ensure the LLM receives the most relevant and helpful in-context example conversations from our gold dataset, we embedded the persona descriptions with sentence transformers \cite{reimers-gurevych-2019-sentence} and identified the most similar $k=2$ examples in the gold dataset.


\begin{table}[t]
    \centering
    \scriptsize
    \tt
    \begin{tabular}{|p{0.45\textwidth}|}
    \hline
         \textcolor{cadmiumgreen}{\#Few-shot examples}\\
         \textcolor{blue}{Player Persona:} Reeves is a teenager and intermediate level Minecraft player. He seeks the NPC's help in finding an ender dragon egg. Reeves enjoys making jokes about the NPCs as he journeys through the game. He hopes this will make them happy and entice them into supporting him.\\
         \textcolor{blue}{Player:} Hey, what can you do for me?\\
         ...\\
         \textcolor{red}{Hillary:} Finally, go through the portal and defeat the ender dragon. The egg will spawn once it's defeated. But you need some equipment. Gear up and take some friends, it's going to be a tough fight.\\
        \textcolor{blue}{Player:} Fine, let me think about it. Thanks for everything! Good luck with your project!\\
        \textcolor{red}{Hillary:} Likewiseee!\\
        \textcolor{blue}{Player:} \textcolor{orange}{***end conversation***}\\
        \textcolor{cadmiumgreen}{\#Instructions}\\
        Respond as if you are the player with the persona below.\\
        \textcolor{blue}{Player Persona:} Trent has played Minecraft at a high level. Trent has seen all there is in the game, or so he think. He is reckless and boisterous and finds a challenge humorous.\\
        \textcolor{blue}{Player:} \\
    \hline
    \end{tabular}
    \caption{Player prompt example.}
    \label{tab:player_utterance_prompts}
\end{table}

\subsubsection{Utterance Generation}
\label{sec:utterances}

In all conversations in our dataset, the player initiates and concludes the conversation. We start by generating the first utterance conditioned on the player persona. The prompt, exemplified in Table~\ref{tab:player_utterance_prompts}, consists of the few-shot example conversations obtained in Sec~\ref{sec:relevant_conversations}, followed by instructions to generate the player's utterance and the player persona. The prompt ends with  
 ``Player:'' to guide the model to generate the utterance. To ensure the model understands which utterances are generated by the player, we replace the player utterance tag (e.g., ``Dan:'' in Figure~\ref{fig:example_conversation}) with ``Player:''.  We iteratively generate the player's utterance and the NPC utterance. In subsequent turns, we include the conversation history. The conversation ends if on the player's turn the LM generated \texttt{``***end conversation***''}.  
 Alternatively, we stop generating the current utterance if the LLM generated ``Function call:'' or ``NPC:''. 

The NPC utterances are generated symmetrically, conditioned on the few-shot examples, NPC persona description, conversation history, and with ``NPC:'' ending the prompt. See example prompt in Appendix~\ref{appendix.npc_agent}. We stop generating the current utterance if the model generated ``Player:'' or ``Function call:''. We set the temperature to 0.9 when generating responses for both the NPC and the player.

\subsubsection{Function call Generation}
\label{sec:function_calls_gen}

Although function calls appear in the middle of the conversation, we generate them as a post-processing step that inserts function calls between player and NPC utterances. This approach allows us to generate accurate function calls without compromising the natural conversation flow. 

Appendix~\ref{appendix.function_calls} provides an example prompt for generating function calls. Unlike previous steps, we provide the model with a small window of the 5 previously logged utterances from the conversation history (including function calls) 
and a placeholder for the target function call inbetween the last player and NPC utterances. 
The instructions include 26 examples of function calls from the gold dataset.  
We stop the generation if the model generated ``Player:'' or ``NPC:''. In contrast to utterance generation, we specifically set the temperature to 0 in order to constrain the LLM to generate one of our predefined function calls.

\ourdataset{} comprises 250 NPC personas, 750 player personas, 49 human-written conversations and 220 automatically-generated conversations. The average length of conversations is 15 turns (excluding function calls), and the average utterance length is 21 words. The player personas, on average, consist of 41 words, while the NPC personas have an average length of around 42 words. 
\section{Evaluation}
\label{sec:evaluation}

To evaluate the quality of the automatically-generated conversations, we asked 3 expert annotators to rate the following aspects of conversations on a scale of 1 to 5 (a higher rating indicating better quality):
\begin{enumerate}[leftmargin=*,itemsep=0em,label=\protect\mycirc{\arabic*}]
    \item \textbf{Player Persona:} The conversation is effective in representing the player's characteristics and behaviors.
    \item \textbf{NPC Persona:} The conversation is effective in representing the NPC characteristics and behaviors.
    \item \textbf{Function Calls:} The function calls are appropriate and adequately support the flow and progression of the conversation.
    \item \textbf{Overall Conversation Flow and Quality:} The conversation is coherent and the dialogue is meaningful.
\end{enumerate}

\begin{table}[t]
    \centering
    \small
    \begin{tabular}{cccc}
        \toprule
        \mycirc{1} & \mycirc{2} & \mycirc{3} & \mycirc{4} \\
        \midrule
         4.08 & 4.30 & 4.21 & 4.13 \\
         \bottomrule
    \end{tabular}
    \caption{Human evaluation results for the automatically-generated conversations.}
    \label{tab:human_eval}
\end{table}

We sampled 60 conversations. Each annotator evaluated 30 conversations, with a 15 conversation overlap between every pair of annotators. The average pairwise Cohen's kappa inter-annotator agreement was 0.42, and the average absolute differences between scores given by multiple annotators were 0.96 for player persona, 0.6 for NPC persona, 0.83 for function calls, and 1.13 for overall conversation flow and quality. 
Table~\ref{tab:human_eval} shows the average scores across conversations. The high scores across criteria indicate that the conversations in \ourdataset{} are coherent and adequately represent the persona, and the use of function calls is appropriate.

\section{Conclusion}
\label{sec:conclusion}

We introduced \ourdataset{}, a persona-driven dialogue dataset for Minecraft. Each example in the dataset includes complex human-written player and NPC personas and a conversation between them. These conversations incorporate natural language utterances, as well as canonical function calls representing game APIs. We collected 49 conversations from experts, and proposed an LM-based approach to extend the dataset, generating over 200 additional conversations. Human evaluation confirmed that the overall quality of the generated conversations is high, that they are consistent with the NPC and player personas, and that function calls were used appropriately. We believe that the \ourdataset{} dataset and the proposed approach offer valuable resources for dialogue system research in gaming. They enable the creation of immersive interactions between players and NPCs, enhancing user experiences in Minecraft and beyond.
\section*{Limitations} 

One limitation of our dataset is the relatively small number (49) of human-written conversations. This limited size may impact the diversity of the dataset, potentially constraining the coverage of various conversation topics and scenarios. Additionally, there is an imbalance between the number of automatically generated conversations (220 instances) and the human-written conversations. This imbalance may introduce biases and discrepancies in the quality and naturalness of the generated conversations compared to the human-written counterparts. To address this limitation, we have taken steps to mitigate the impact by releasing the human-written conversations separately. This allows researchers and developers to have access to these high-quality human-written instances, enabling them to make informed decisions when using our dataset.

\section*{Ethical Statement}

\paragraph{Bias.} A big part of our dataset is generated by LMs. It is important to note that LMs have the potential to generate biased, offensive or harmful content in response to certain prompts. In order to mitigate this risk, the authors of this paper carefully evaluated the narratives to ensure they are free from such content. The authors also thoroughly reviewed the written personas and conversations to ensure the absence of offensive content and sensitive information, and guarantee their adherence solely to the Minecraft game context.

\paragraph{Compensation.} The expert annotators were compensated at an hourly rate of 20 USD, which exceeds the minimum wage in the United States. The human evaluation was conducted by the authors of this paper. 

\section*{Acknowledgments}

This work was funded, in part, by Microsoft, the Vector Institute for AI, Canada CIFAR AI Chairs program, Accelerate Foundation Models Research Program
Award from Microsoft, an NSERC discovery grant, and a research gift from AI2.

\bibliography{custom,anthology}
\bibliographystyle{acl_natbib}

\appendix

\section{Appendices}

\subsection{NPC Utterance Prompt}
\label{appendix.npc_agent}

Table~\ref{tab:npc_utterance_prompts} shows an example prompt for generating the NPC's responses.


\begin{table}[t]
    \centering
    \scriptsize
    \tt
    \begin{tabular}{|p{0.45\textwidth}|}
    \hline
         \textcolor{cadmiumgreen}{\#Few-shot examples}\\
         \textcolor{blue}{Player Persona:} Hillary is a bot in Minecraft. His speciality is armor and farming. He encourages players to trade with similar NPCs who craft armor and is a fan of teamwork. He however requires appreciation in the form of the rate Heart of the Sea.\\
         \textcolor{blue}{Reeves:} Hey, what can you do for me?\\
         ...\\
         \textcolor{red}{NPC:} Finally, go through the portal and defeat the ender dragon. The egg will spawn once it's defeated. But you need some equipment. Gear up and take some friends, it's going to be a tough fight.\\
        \textcolor{blue}{Reeves:} Fine, let me think about it. Thanks for everything! Good luck with your project!\\
        \textcolor{red}{NPC:} Likewiseee!\\
        \textcolor{blue}{Reeves:} \textcolor{orange}{***end conversation***}\\
        \textcolor{cadmiumgreen}{\#Instructions}\\
        Respond as if you are the NPC with the persona below.\\
        \textcolor{red}{NPC Persona:} Tabatha is a strange NPC in the woods of Minecraft. They are found near a large fire pit with skeletal remains around. Tabatha speaks in a strange manner and in the third person, both warning players to leave and challenging them to stay at the same time.\\
        \textcolor{red}{NPC:} \\
    \hline
    \end{tabular}
    \caption{NPC prompt example.}
    \label{tab:npc_utterance_prompts}
\end{table}

\subsection{Function Call Generation Prompt}
\label{appendix.function_calls}

Table~\ref{tab:function_call_example} shows an example function call generation prompt. 


\begin{table}[t]
    \centering
    \scriptsize
    \tt
    \begin{tabular}{|p{0.45\textwidth}|}
    \hline
         \textcolor{cadmiumgreen}{\#Instructions}\\
         You may only answer ``Call chat'' or use answers that start with one or a combination of the following function calls: Call get crafting recipe on, Call locate a resource on, Call locate a place on, Call follow the player, Call come to the player, Call mine a block on, Call get count of item in inventory on, Call list all items in inventory, Call drop item on, Call close a chest, Call open a chest, Call take items from the chest on, Call put items into the chest on, Call locate a block on, Call take items from player on\\         
         \textcolor{cadmiumgreen}{\#Few-shot examples and input}\\
         \textcolor{blue}{Player:} This chest is locked.\\
         \textcolor{orange}{Function call:} Call open a chest\\
         \textcolor{red}{NPC:} Silly me, let me unlock it for you!\\
         \textcolor{blue}{Player:} I would like these torches and some of the quartz, please.\\
         \textcolor{orange}{Function call:} \_\\
         \textcolor{red}{NPC:} Here you are, a stack of torches and half of the quartz. Oh no, it appears your inventory is full. Would you like me to carry these materials for you while you build the bridge?\\
        \textcolor{cadmiumgreen}{\#Instructions}\\
        Replace \_ with the proper Function call.\\
        Answer: \\
    \hline
    \end{tabular}
    \caption{Function call prompt example. The answer to the given example is ``Call take items from the chest on torches and quartz''.}
    \label{tab:function_call_example}
\end{table}

\subsection{List of all Function Calls}
\label{appendix.all_function_calls}

Table~\ref{tab:all_function_calls} shows all the function calls in \ourdataset{}.

\begin{table}[t]
    \centering
    \scriptsize
    \begin{tabularx}{\columnwidth}{Xl}
        \toprule
        \textbf{Function Call} & \textbf{Args}\\
        \midrule
         Call chat & N/A \\
         Call come to the player & N/A \\
         Call list all items in inventory & N/A  \\
         Call get crafting recipe on & iron sword, etc. \\
         Call locate a place on & realms, caves, etc. \\
         Call check if item is in inventory on & pickaxe, sword etc.\\
         Call drop item on &potions, sword, etc.\\
         Call take items from player on & diamonds, etc.\\
         Call locate a resource on & diamonds, etc. \\
         Call follow the player & N/A \\
         Call mine a block on & Iron ore, etc. \\
         Call craft an item on & bow, etc. \\
         Call get count of item in inventory on & gold, etc. \\
         Call close a chest & N/A \\
         Call open a chest & N/A \\
         Call take items from the chest on & armor, etc. \\
         Call put items into the chest on & diamond, etc. \\
         Call locate a block on & iron ore, etc \\
         Call End Conversation & N/A \\
         Call get mining recipe on & cobblestone, etc.\\
         \bottomrule
    \end{tabularx}
    \caption{List of all function calls from \ourdataset{}.}
    \label{tab:all_function_calls}
\end{table}

\end{document}